\documentclass[final,5p,times,twocolumn,authoryear]{elsarticle}
\usepackage{float}
\usepackage{amssymb}
\usepackage{amsmath}
\usepackage{xcolor}
\usepackage{booktabs}
\usepackage[caption=false,font=normalsize,labelfont=sf,textfont=sf]{subfig}
\usepackage{url}

\journal{Accident Analysis and Prevention}

\begin{document}

\begin{frontmatter}

\title{Practical validation of synthetic pre-crash scenarios} 

\author[a,b]{Jian Wu\corref{cor1}}
\cortext[cor1]{Corresponding author}
\ead{jian.wu@chalmers.se}

\author[a]{Ulrich Sander}
\author[b,c]{Carol Flannagan}
\author[b]{Jonas Bärgman}

\affiliation[a]{organization={Safety Center, Volvo Cars},
            addressline={41878 Göteborg, Sweden}
            }

\affiliation[b]{organization={Department of Mechanics and Maritime Sciences, Chalmers University of Technology},
            addressline={41756 Göteborg, Sweden}
            }

\affiliation[c]{organization={University of Michigan Transportation Research Institute},
            addressline={Ann Arbor, Michigan 48109, USA}
            }

\begin{abstract}
The representativeness of synthetic pre-crash scenarios is crucial for assessing the safety impact of Driving Automation Systems through virtual simulations.
However, a gap remains in the robust evaluation of synthetic pre-crash scenarios' practical equivalence to their real-world counterparts; that is, whether they are similar enough for the intended assessment purpose.
Conventional significance testing is inadequate, as it focuses on detecting differences rather than establishing practical equivalence.
This study addresses the research gap by extending our previous work on a Bayesian Region of Practical Equivalence (ROPE)-based equivalence testing framework by introducing a binning-based approach to define appropriate statistics and equivalence criteria.
Two binning-based statistics are proposed to measure practically meaningful distributional differences between datasets in the context of safety impact assessment.
The framework's applicability is demonstrated through a case study, which tests the practical equivalence of two synthetic rear-end pre-crash datasets with a previously developed reference dataset in the context of the safety impact assessment of an Automatic Emergency Braking system.
The results show that the framework provides informative quantitative assessments of practical equivalence as well as diagnostic insights into the divergence of datasets.
Although the demonstration focuses on rear-end pre-crash scenarios, the framework is generic and extensible to broader validation contexts, providing an interpretable and principled basis for practical equivalence assessment across diverse synthetic data applications.
\end{abstract}




\end{frontmatter}

\section{Introduction} \label{section:intro}
Driving Automation Systems (DAS) \citep{sae_j3016_2021}, including Advanced Driver Assistance Systems (ADAS) and Automated Driving Systems (ADS), are expected to reduce crash risk and improve traffic safety \citep{pradhan2022impact}.
To assess their safety impact, \emph{virtual safety assessment} has become a primary procedure due to its low cost and high efficiency compared to conventional field tests \citep{dona2022virtual, cai2022survey, szalay2023critical, wu2025model}.
In this virtual paradigm, pre-crash scenarios—short time sequences describing driver, vehicle, and environmental dynamics leading up to a potential collision—are simulated in two conditions: baseline (without the DAS under assessment) and treatment (with the DAS).
These pre-crash scenarios must align with the assessment objective and should encompass all relevant elements that may impact the performance of the technology under assessment \citep{wimmer2023harmonized}.
The safety effects of a system can be estimated by comparing the outcomes of the simulated baseline and treatment scenarios \citep{bargman2017counterfactual, baron2020repeatable, szalay2023critical, wimmer2023harmonized}.

The assessment procedure requires that the pre-crash scenarios used are adequate and accurately represent real-world conditions, to ensure a statistically sound and unbiased comparison \citep{dona2022virtual, cai2022survey, wimmer2023harmonized, wu2024generation}.
However, real-world pre-crash data are typically limited in quantity and suffer from sampling bias and coverage issues.
For example, naturalistic driving studies, such as the Second Strategic Highway Research Program (SHRP2) Naturalistic Driving Study (NDS), provide extensive exposure to everyday driving behavior, recording millions of kilometers of real-world traffic interactions \citep{hankey2016description}.
However, despite their large scale, these studies capture relatively few crashes, particularly high-severity ones, over-representing low-severity crashes \citep{bargman2017counterfactual, wu2025model}.
In contrast, in-depth crash databases, such as the Crash Investigation Sampling System (CISS) in the United States and the German In-Depth Accident Study (GIDAS) Pre-Crash Matrix (PCM), provide detailed recorded or reconstructed kinematic data and associated behavioral and environmental factors, but represent only a small subset of crashes; high-severity crashes are over-represented \citep{schubert2017gidas, zhang2019crash}.

To overcome these limitations, a common strategy is to create synthetic pre-crash scenarios using statistical and/or behavioral models derived from real-world data \citep{scanlon2021waymo, gambi2019generating, wang2022autonomous, wu2025model}.
This approach ensures the generation of a sufficient number of scenarios.
However, a critical gap remains in validating the representativeness of the resulting synthetic scenarios.
In particular, it is especially important to determine whether the synthetic scenarios are practically equivalent to their real-world counterparts for the intended assessment, meaning they are similar enough that any remaining differences are negligible in practice.
Without proper validation, biases such as the over-representation of severe crashes may remain undetected, leading to misleading or biased assessments \citep{olleja2022can, bargman2017counterfactual, hamdane2015issues}.

Conventional statistical significance tests are commonly used as formal validation tools in traffic safety research \citep{lord2010statistical, daamen2014traffic}.
However, these methods are inherently difference-oriented: they are designed to detect differences rather than to establish equivalence \citep{anderson2000null}.
Specifically, the absence of a statistically significant difference does not imply equivalence, nor does a statistically significant difference necessarily indicate a practically meaningful discrepancy in a given application context \citep{gibbs2013errors, wu2025practical}.

To address this gap, a promising solution lies in practical equivalence testing.
It is a set of statistical approaches aimed at determining whether two treatments, processes, or groups are sufficiently similar that any observed differences are small enough to be ignored in practice \citep{limentani2005beyond, greene2000claims}.
While such methods are well established in fields such as medicine and psychology \citep{lakens2018equivalence}, their adoption in traffic safety and automated driving research remains limited to date.

\begin{figure}[!t]
    \centering
    \subfloat[]{\includegraphics[width=0.14\textwidth]{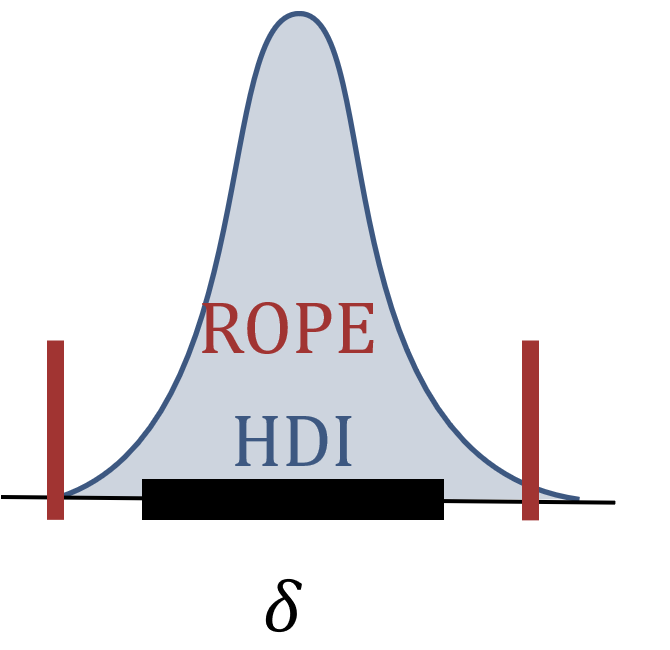}}
    \hfil
    \subfloat[]{\includegraphics[width=0.17\textwidth]{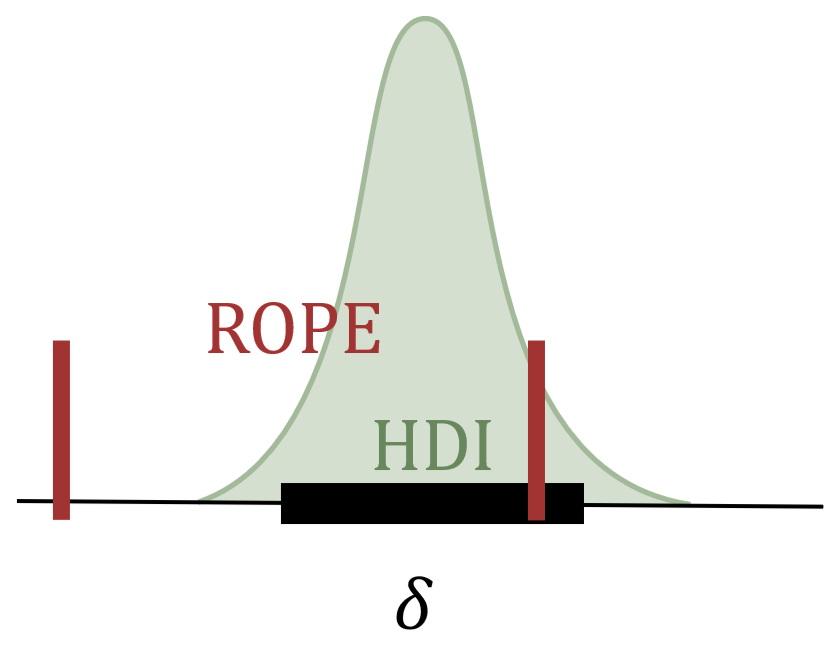}}
    \caption{Illustration of the ROPE concept in Bayesian equivalence testing: (a) equivalence and (b) non-equivalence.
    The shaded curves represent the posterior distributions of a parameter or statistic, $\delta$.
    Red vertical lines denote the ROPE boundaries, and black bars indicate the HDIs.
    Equivalence is supported when the HDI lies entirely within the ROPE.}
    \label{fig:rope_concept}
\end{figure}

Among these methods, Bayesian approaches based on the Region of Practical Equivalence (ROPE) \citep{schwaferts2020bayesian} offer a straightforward interpretation of similarity and naturally integrate domain-specific thresholds \citep{kruschke2018rejecting, wu2025practical}, making them particularly well-suited in virtual safety impact assessments of DAS.
Rather than asking whether two samples originate from exactly the same distribution, ROPE-based methods assess whether the highest density intervals (HDIs) of posterior statistic estimates—such as the difference between means or variances—fall entirely within a predefined region of values considered practically equivalent.
Equivalence is quantified as the probability that a parameter falls within a region of negligible difference, thereby providing an interpretable probabilistic statement of similarity that also integrates prior information.
The underlying logic of this procedure is illustrated in Fig.~\ref{fig:rope_concept}, where the relative positions of the HDI and ROPE determine whether practical equivalence can be concluded.

\begin{figure}[!t]
    \centering
    \includegraphics[width=1\linewidth]{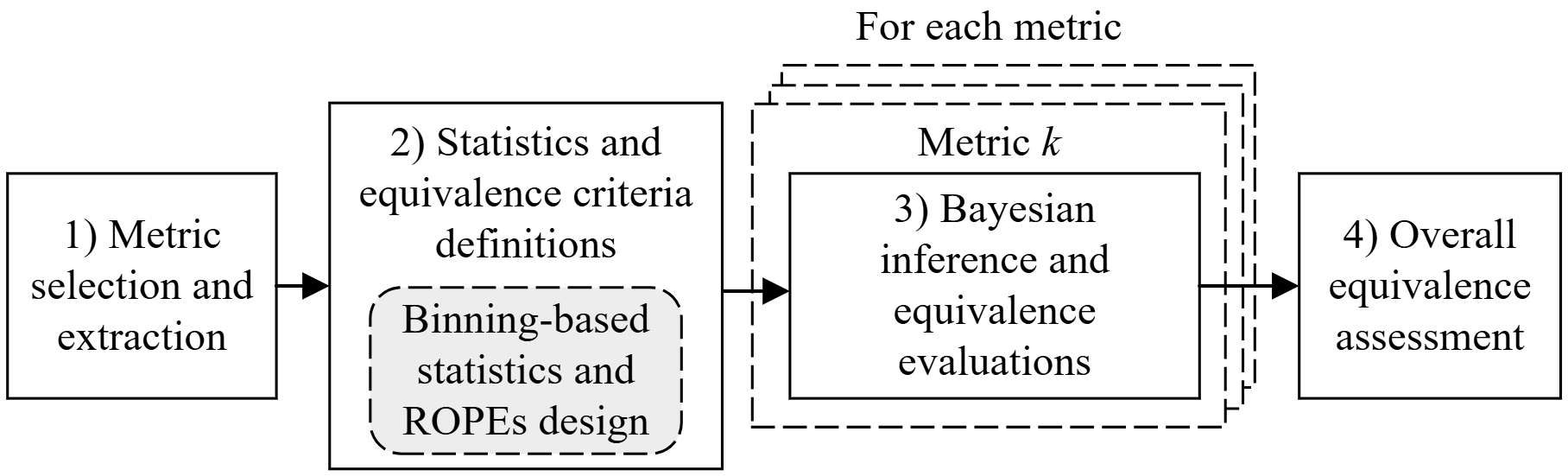}
    \caption{Flowchart of the proposed equivalence testing method.
    The gray component in Step 2 represents the part newly developed in this work.}
    \label{fig:methodolgy}
\end{figure}

Our earlier work proposed a Bayesian ROPE-based equivalence test framework (shown in Fig. \ref{fig:methodolgy}) for validating synthetic pre-crash scenarios \citep{wu2025practical}.
Step 2 of the framework is intentionally flexible, allowing users to define the statistics and equivalence criteria based on the specific needs of the assessment.

This study aims to provide clear and practical guidelines for implementing this step by introducing a novel approach that uses binning-based statistics and ROPEs designed in a transparent and interpretable manner (illustrated by the gray component in Fig.~\ref{fig:methodolgy}).
\textcolor{black}{The approach is illustrated through a case study, which tests the practical equivalence of two synthetic rear-end pre-crash datasets with a previously developed reference dataset \citep{wu2025model} in the context of the safety impact assessment for an Automatic Emergency Braking (AEB) system.}
The case study demonstrates how the framework can support both quantitative assessments of practical similarity and diagnostic analysis, such as identifying which regions of the data contribute most to non-equivalence.
It also illustrates how sample weighting can mitigate sampling bias within a defined assessment scope, although its effectiveness depends on the underlying data structure and must be evaluated on a case-by-case basis.

Although rear-end pre-crashes are used in the case study, the validation framework is inherently generic and can be readily extended to contexts beyond safety impact assessments, such as model validation in traffic flow simulation, comparative evaluation of behavioral models, or assessment of synthetic data used for system testing.
By providing an interpretable and probabilistically sound foundation for assessing practical equivalence, the framework is a flexible tool for the systematic validation of synthetic data against a reference.

\section{Methodology} \label{section:method}
This section describes the extended Bayesian ROPE-based practical equivalence testing framework.
Its primary objective is to assess whether a synthetic pre-crash dataset is practically equivalent to a representative reference dataset for an intended safety impact assessment.

\subsection{Proposed Equivalence Testing Method} \label{section:method_steps}
The proposed equivalence testing method contains the four steps shown in Fig. \ref{fig:methodolgy}.
Pre-crash data typically consist of time series outlining the dynamics and trajectories of (at a minimum) the road users involved in the crash.
Because direct comparison of time-series data often is impractical, they are characterized by a set of derived variables, referred to here as ``metrics''.
In this study, metrics are the quantitative variables used to characterize a scenario for equivalence testing.
These may include scenario-descriptive parameters such as initial speeds and relative positions, obtained either directly from the time series or through a parameterization of the scenario, as well as outcome-based quantities such as Delta-v or estimated injury risk (when relevant to the safety impact assessment).
The framework does not depend on a specific parameterization; it accommodates any set of metrics deemed meaningful for evaluating representativeness in the intended assessment context.

\subsubsection{Step 1: Metric selection and extraction} \label{section:method_steps_1}
This step involves selecting the metrics most relevant to the intended assessment and extracting or deriving them from both the reference and synthetic datasets.
The reason for selecting only the most relevant metrics is that, while using more metrics can make the test more comprehensive, it also heightens the risk of making a false non-equivalence error.
In such cases, non-equivalence in less relevant metrics may dominate the validation outcome and lead to an erroneous rejection of an otherwise representative dataset.
Therefore, the assessment only considers the selected metrics.

\subsubsection{Step 2: Statistics and equivalence criteria definitions} \label{section:method_steps_2}
For each metric, two statistics, $\theta$ and $\Theta$, are defined to assess the differences between its distributions in the two compared datasets.
(More details about these statistics will be provided in Section \ref{section:statistics}.)

Subsequently, practical equivalence criteria are established for the statistics by specifying both the posterior probability threshold ($\alpha$) and ROPEs ($\left[0, \theta_{\mathrm{thd}}\right]$ and $\left[0, \Theta_{\mathrm{thd}}\right]$, where $\theta_{\mathrm{thd}}$ and $\Theta_{\mathrm{thd}}$ are ROPE thresholds).
Typically, a 95\% posterior probability threshold is selected, and the ROPE is determined based on expert judgment and domain-specific knowledge relevant to the application.
For a metric to be deemed practically equivalent between datasets, the 95\% HDIs of both statistics' posterior distributions must lie entirely within the defined ROPEs.

\subsubsection{Step 3: Bayesian inference and equivalence evaluations} \label{section:method_step3}
For each metric, a set of Bayesian distribution models (e.g., exponential, normal, log-normal, gamma, or mixture models) is chosen based on expert judgment.
The models are fitted separately to empirical distributions of the metric in reference and synthetic datasets.
If the datasets are weighted, sample weights are incorporated directly into the Bayesian model through a weighted likelihood formulation.
Specifically, each observation contributes to the total log-likelihood in proportion to its assigned weight, ensuring that the posterior distribution reflects the representativeness of the weighted data.
Leave-one-out cross-validation \citep{vehtari2017practical} is employed to select the optimal distribution model for each dataset.

Posterior samples, each representing a draw of the predictive distribution parameters, are obtained from the optimal Bayesian models fitted to the two datasets.
The two binning-based statistics $\theta$ and $\Theta$ are computed for each paired set of posterior samples from both distribution models.
Finally, the HDI of each statistic's posterior distribution is evaluated against the practical equivalence criteria defined in Step 2.
If each HDI lies entirely within its corresponding ROPE, the data support the practical equivalence of the metric between the compared datasets.
\textcolor{black}{The Python implementation of the Bayesian distribution fitting and binning-based statistics computation functions used in this study is publicly available in the \texttt{bayes-binned-equivalence} repository \citep{Wu_bayes-binned-equivalence_2026}.}

\subsubsection{Step 4: Overall equivalence assessment}
In the previous steps, multiple equivalence tests have been conducted for the selected metrics.
This step aims to establish clear overall equivalence criteria that synthesize these individual results into a single conclusion.
The criteria should effectively integrate the outcomes of individual tests with expert-based weighting of the relative importance of different metrics, ensuring that the synthetic dataset adequately represents the real-world data in the aspects most relevant to the intended assessment.

In theory, two datasets are considered equivalent only if all metrics individually demonstrate equivalence.
However, in practice, less stringent approaches may be justified.
For instance, overall equivalence may still be concluded if equivalence is achieved for a subset of the selected metrics—specifically those identified as most critical within the already relevance-filtered set—provided that the remaining less-critical metrics do not exhibit substantial deviations.

Two additional considerations are important here.
First, the overall equivalence criteria should be clearly defined in advance.
If applicable, the definitions should also include a quantitative explanation of what constitutes a ``substantial deviation'' for less-critical metrics.
Importantly, the rationale for adopting less stringent criteria should be based on the trade-off between strictness and inclusiveness.
As the number of evaluation metrics increases, so does the likelihood that at least one of them will fail to meet the equivalence criteria.
Allowing some flexibility for less critical metrics enables a more holistic and practically meaningful judgment of equivalence, supported by a transparent understanding of where and how the synthetic data diverges from the reference data.

Second, it is essential to emphasize that practical equivalence testing relies heavily on expert judgment and reasoning, rather than purely numerical outcomes.
It is generally not possible to define practical equivalence without making decisions based on some level of expertise.
Therefore, the rationale for declaring equivalence or non-equivalence must be thoroughly documented, including the motivations behind the selection of metrics, statistics, ROPEs, and overall equivalence criteria.

\subsection{Two Binning-Based Statistics} \label{section:statistics}
The two proposed statistics, $\theta$ and $\Theta$, are designed to quantify practical differences between the (one-dimensional) reference and synthetic distributions within a specific safety impact assessment context.
These statistics are derived through a binning process that strikes a balance between interpretability and robustness.
Interpretability is achieved by decomposing distributional differences into deviations across explicit, localized regions of the metric’s range, which can be directly related to assessment-relevant regimes.
Robustness is achieved by aggregating data within bins, thereby reducing sensitivity to small-scale noise (random fluctuations that do not carry meaningful information for the assessment) while still capturing systematic differences (consistent and structured deviations across the distribution that reflect genuine discrepancies between the datasets).

The reference distribution is first partitioned into $N$ bins, each containing (approximately) the same proportion of the total data.
The same bin boundaries are then applied to the synthetic distribution, resulting in paired bin proportions that enable a direct, one-to-one comparison between the two distributions.

It is important to note that not all bin groups are equally relevant for the intended safety impact assessment.
The practical relevance of each bin depends on the specific objective of the assessment.
For instance, under many safety-impact objectives, bins corresponding to safety-critical ranges of a metric or to conditions in which the assessed system performs poorly should have a greater influence on the overall assessment outcome.
Bin weights can reflect this heterogeneity in practical relevance.
They link statistical deviations across bins with their potential impact on assessment results, thereby bridging the gap between purely statistical differences and practically meaningful consequences.
Bin weighting thus constitutes a key component of the proposed method.
Details regarding the bin weighting design and implementation are provided in Section \ref{section:bin_weight_design}.

The two proposed statistics, $\theta$ and $\Theta$, integrate the weighted deviations across bins to quantify the overall degree of practical difference between the reference and synthetic distributions.
Let $\omega_i$ denote the weight of the $i$-th bin and $P_{\mathrm{ref},i}$ and $P_{\mathrm{syn},i}$ denote the proportions of data in the $i$-th bin for the reference and synthetic distributions, respectively.
$\Delta P_i (= P_{\mathrm{syn},i} - P_{\mathrm{ref},i})$ denotes the proportion difference for the $i$-th bin between two distributions.
Then, the two statistics $\theta$ and $\Theta$ are defined as:
\begin{equation} \label{eq:theta}
    \theta = \max_{1 \leq i \leq N} \left(\left|\frac{\Delta P_i}{P_{\mathrm{ref},i}}\right| \cdot \omega_i \right),
\end{equation}
\begin{equation} \label{eq:Theta}
    \Theta = \sum_{i = 1}^{N} \left| \Delta P_i \right| \cdot \omega_i.
\end{equation}
The maximum weighted absolute relative deviation across bins (i.e., a worst-case perspective) is captured by $\theta$, whereas $\Theta$ represents the total weighted absolute deviation (i.e., an aggregate perspective).
Together, they provide complementary views of the practical distributional difference between the synthetic and reference data.

\begin{figure}[ht]
    \centering
    \includegraphics[width=.7\linewidth]{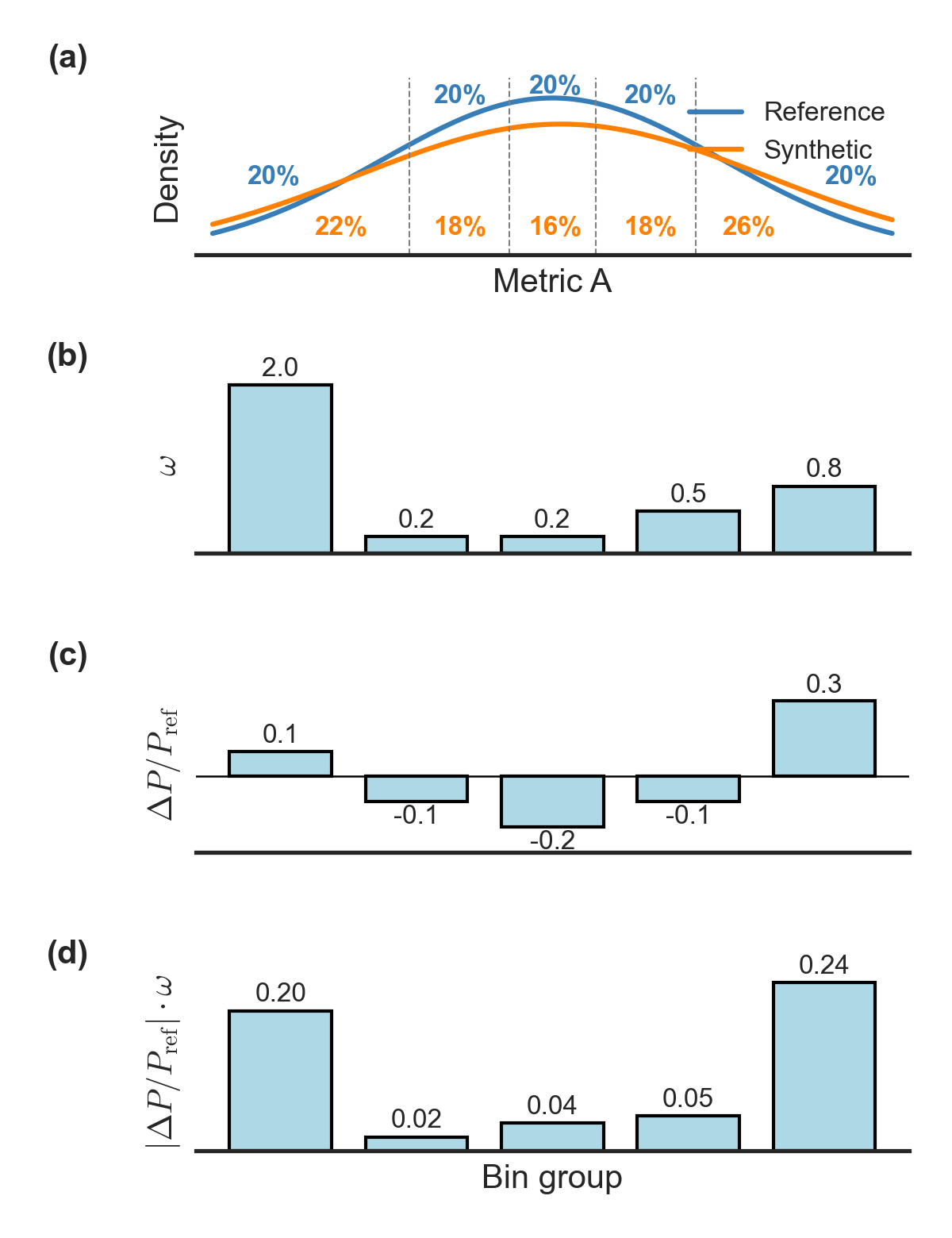}
    \caption{Illustration of the procedure to compute $\theta$.}
    \label{fig:binning_process}
\end{figure}

An illustration of the procedure to compute $\theta$ for the reference and synthetic distributions is shown in Fig.~\ref{fig:binning_process}.
\begin{enumerate}
    \item Fig.~\ref{fig:binning_process}(a) illustrates the binning process used in the computation of $\theta$ and $\Theta$.
    The reference distribution (blue) is divided into five quantile-based bins, each containing 20\% of the reference data.
    The same bin boundaries are then applied to the synthetic distribution (orange), resulting in bin proportions of 22\%, 18\%, 16\%, 18\%, and 26\%.
    \item The weights $\omega$ (illustrated in Fig.~\ref{fig:binning_process}(b) as 2.0, 0.2, 0.2, 0.5, and 0.8) represent the relevance of each bin to the intended assessment.
    The method for setting these weights is described in Section \ref{section:bin_weight_design}.
    \item The relative deviations $\Delta P/P_\mathrm{ref}$ (illustrated in Fig.~\ref{fig:binning_process}(c) as 0.1, -0.1, -0.2, -0.1, and 0.3) measure the divergence between the compared bin pairs in the reference and synthetic distributions.
    \item The weighted absolute relative deviations $|\Delta P/P_\mathrm{ref}|\cdot \omega$ (illustrated in Fig.~\ref{fig:binning_process}(d) as 0.20, 0.02, 0.04, 0.05, and 0.24) then scale the statistical divergence by the practical relevance of each bin, represented by the corresponding weight illustrated in Fig.~\ref{fig:binning_process}(b), emphasizing differences in high-weight regions.
    In this example, the last bin shows the maximum weighted absolute relative deviation, which determines the value of $\theta$.
    Accordingly, $\theta = 0.24$.
\end{enumerate}

It is essential to note that the bin boundaries and bin weights are intrinsic components of the definitions of $\theta$ and $\Theta$.
As described in Section \ref{section:method_step3}, these statistics are computed for each paired set of posterior samples from the Bayesian models fitted to the two datasets being compared.
Consequently, for every posterior draw, the bin boundaries and the corresponding bin weights are recalculated based on the posterior samples from the reference distribution model.

\subsection{Bin Weights} \label{section:bin_weight_design}
We propose an empirical procedure for determining bin weights:
\begin{enumerate}
    \item Re-simulate the reference pre-crash scenarios with a virtual representation of the DAS under assessment to generate system-specific outcome data for each scenario.
    The result may be a sparse distribution due to the limited size of the reference dataset.
    \item Identify a set of re-simulation outcome measures most relevant to the assessment purpose, such as residual impact speed or estimated injury risk, denoted as $[x_1, x_2, \dots, x_l]$.
    \item Define a weight function $f(X)$ that assigns a bin weight based on these outcome measures:
    \begin{equation}
        \omega_i = f(X_i),
    \end{equation}
    where $\omega_i$ is the weight of the $i$-th bin, and $X_i = [x_1^{(i)}, x_2^{(i)}, \dots, x_l^{(i)}]$ represents the selected outcome measures for the $i$-th bin.
\end{enumerate}

The choice of outcome measures and the design of $f(X)$ should be guided by domain knowledge and expert judgment in addition to the assessment objective, to ensure that the resulting bin weights accurately capture the practical relevance of the corresponding bins.
\textcolor{black}{By combining statistical deviations between datasets with bin-level weights that reflect their importance within the assessment scope, the two statistics effectively quantify the practical differences between the compared datasets for the intended assessment.}
An example of the bin-weight function is presented in Section \ref{section:bin_weight}.

\subsection{Parameter Settings} \label{section:param_setting}
The parameters  $N$, $\alpha$, $\theta_\mathrm{thd}$, and $\Theta_\mathrm{thd}$, described below, influence the behavior and sensitivity of the proposed testing method.

The choice of the number of bins $N$ is particularly critical, since it directly determines the granularity of the comparison between the reference and synthetic distributions, and consequently affects the evaluation of the two statistics $\theta$ and $\Theta$.
A larger $N$ yields a more stringent test by enabling finer-grained comparisons, thereby improving the ability to detect localized differences.
However, increasing $N$ also heightens the sensitivity to variance and amplifies the cumulative effect of multiple comparisons.
In practical terms, using more bins increases the likelihood that at least one bin will fail to meet the equivalence criteria, and thus that the overall test will conclude the datasets are not equivalent.
Conversely, a smaller $N$ reduces noise but may fail to catch localized distributional differences.

To balance robustness and granularity, we adopt a simple, practical guideline for selecting $N$: each bin should contain a sufficient number of reference data points to ensure stable proportion estimates and reliable equivalence evaluation.
Further, the number of bins should not exceed some predefined threshold to avoid the cumulative effect of too many comparisons.
Accordingly, the number of bins is determined as:
\begin{equation}
    N = \text{min} \left( \lfloor \frac{n}{m} \rfloor, N_\mathrm{max} \right),
\end{equation}
where $n$ is the number of reference samples, $m$ is the desired minimum number of samples per bin, and $N_\mathrm{max}$ is the maximum allowable number of bins.
Following conventional statistical recommendations based on the Central Limit Theorem, we suggest $m \in [30, 50]$.
To control test sensitivity and mitigate multiple-comparisons effects, we recommend capping the number of bins at $N_\mathrm{max} = 20$.

The remaining three parameters $\alpha$, $\theta_\mathrm{thd}$, and $\Theta_\mathrm{thd}$ jointly define the practical equivalence criteria.
Conventionally, $\alpha$ is set to 0.95.
The parameters $\theta_\mathrm{thd}$ and $\Theta_\mathrm{thd}$ are thresholds shaping the ROPEs for the statistics $\theta$ and $\Theta$, respectively.
These thresholds should be set based on expert judgment to reflect the relevance and priorities of the intended assessment.
A practical example is provided in Section \ref{section:parameters}.

It is important to note that a common pair of ROPE thresholds can be applied across different metrics.
This is possible because, although the bins are partitioned separately based on the reference distribution of each metric, the bin weights are consistently determined by the same weight function, ensuring comparability of the two statistics across metrics.

\section{Demonstration}
This study demonstrates the proposed practical equivalence testing method by comparing two rear-end pre-crash scenario datasets with a reference dataset in the context of a safety impact assessment for an AEB system \citep{V4SAFETYRepo2024} developed within the V4SAFETY Project \citep{v4safety}.

\subsection{Datasets} \label{section:datasets}
\subsubsection{Reference dataset}
The first dataset comprises 200 rear-end pre-crash scenarios randomly sampled with replacement from the QUADRIS pre-crash dataset \citep{Wu_QUADRIS_project_pre-crash_near-crash}.
The full QUADRIS dataset contains 5,000 weighted rear-end pre-crash scenarios generated by modeling real-world rear-end pre-crash data and is designed to represent the U.S. rear-end crash population across the full severity range, from minor physical contact to severe injuries and fatalities \citep{wu2025model}.
We consider the QUADRIS dataset to be the most comprehensive representation to date of U.S. rear-end crashes across all severity levels.
Because real-world pre-crash datasets are typically much smaller than the full QUADRIS dataset, a random sample of 200 scenarios was drawn to approximate realistic dataset sizes; this subset is treated as the ``reference'' dataset in the present study.

\subsubsection{PCM dataset}
The second dataset consists of 866 reconstructed concrete rear-end pre-crash scenarios from the GIDAS-PCM dataset \citep{schubert2017gidas, wu2025practical}, initiated in 2011.
Hereafter referred to as the ``PCM'' dataset, it is a dedicated subset of the GIDAS database, which collects on-scene accident investigations involving personal injury in Hannover and Dresden, Germany \citep{schubert2017gidas}.

Unlike the reference dataset, which spans the full severity range, the PCM dataset includes only injury-involved crashes, with lower inclusion probabilities for less severe injuries \citep{wu2025practical}.
To mitigate this known bias, a weighted version of the PCM dataset was created, as detailed in Section \ref{section:results_weighted_pcm}.
Both the raw and weighted PCM datasets were compared with the reference dataset.

\subsubsection{SCM-based dataset}
We also wanted to include a dataset of rear-end pre-crash scenarios generated via traffic simulation, an approach that contrasts with the crash reconstruction approach used to create the PCM dataset \citep{shah2018airsim, baron2020repeatable, fries2022driver}.
Ideally, such a dataset would be generated either by simulating everyday multi-agent traffic and extracting crash events as they naturally occur or by scenario-based simulation, where selected critical scenarios are simulated repeatedly, and the resulting outcomes are weighted by their real-world occurrence frequencies.
However, to our knowledge, no such dataset is currently available.
Therefore, we created one for demonstration purposes, based on the QUADRIS dataset and the Stochastic Cognitive Model (SCM) \citep{fries2022driver, BMW_SCM}.

This dataset consists of 7,888 synthetic rear-end pre-crash scenarios that were obtained by re-simulating each of the 5,000 pre-crash scenarios (hereafter referred to as the ``seed cases'') in the full QUADRIS dataset 2,000 times. 
The lead-vehicle behavior and initial following distance were fixed as in the seed cases, and the following vehicle was governed by the SCM.

The SCM has been under development by BMW and partners since 2014 and is used in the openPASS simulation framework \citep{fries2022driver}.
By modeling cognitive processes, the SCM aims to ensure a realistic representation of the traffic in the simulation across a wide range of scenarios, including possible collision scenarios \citep{BMW_SCM}.
Given the same input, SCM generates various behavioral outcomes by sampling driver perceptions and actions from probabilistic distribution models; thus, the 2,000 re-simulation outcomes of a single seed case are not identical.
Only around 0.08\% of all the simulations resulted in a crash.
Of the 5,000 seed cases, 651 generated at least one re-simulated crash, with the number of crashes per seed case ranging from 0 to 959 out of the 2,000 simulations.

For each seed case, the number of re-simulated crashes depends on the number of simulation runs and the SCM crash probability in that scenario.
Ideally, the number of simulation runs should be chosen—or the resulting crashes weighted—in proportion to the real-world occurrence frequency of the corresponding seed case.
Formally,
\begin{equation} \label{eq:seed_frequency}
    n_{\mathrm{sim,}i} \cdot \omega_{\mathrm{sim,}i} \propto f_i,
\end{equation}
where $n_{\mathrm{sim,}i}$ is the number of simulations conducted for the $i$-th seed case, $\omega_{\mathrm{sim,}i}$ is the weight assigned to all re-simulated crashes originating from that seed case, and $f_i$ denotes the real-world occurrence frequency of the $i$-th seed case scenario.
This condition ensures that the weighted dataset of all re-simulated pre-crash scenarios reflects the true prevalence of seed-case types in real traffic, thereby preventing the over-representation of seed cases that are frequently simulated but rare in reality.

However, the real-world frequencies $f_i$ of the seed case scenarios are not available.
Therefore, we utilize a property of the QUADRIS dataset: the weight of each pre-crash scenario is proportional to the real-world frequency of that crash type.
Formally,
\begin{equation} \label{eq:seed_weight}
    \omega_i \propto f_i \cdot p_{\mathrm{c},i},
\end{equation}
where $\omega_i$ is the weight of the $i$-th seed case in the QUADRIS dataset and $p_{\mathrm{c},i}$ denotes the crash probability of human drivers in that scenario.
Combining (\ref{eq:seed_frequency}) and (\ref{eq:seed_weight}) gives:
\begin{equation}
    n_{\mathrm{sim,}i} \cdot \omega_{\mathrm{sim,}i} \cdot p_{\mathrm{c},i} \propto f_i \cdot p_{\mathrm{c},i} \propto \omega_i.
\end{equation}
If we assume that, in seed cases with at least one re-simulated crash, SCM has the same crash probability as human drivers, that is:
\begin{equation}
    \hat{p}_{\mathrm{c,SCM},i} = p_{\mathrm{c},i} \quad \text{for seed cases with } \hat{p}_{\mathrm{c,SCM},i} > 0,
\end{equation}
then we obtain:
\begin{equation} \label{eq:assumpation}
    n_{\mathrm{sim,}i} \cdot \omega_{\mathrm{sim,}i} \cdot \hat{p}_{\mathrm{c,SCM},i}  \propto \omega_i \quad \text{for } \hat{p}_{\mathrm{c,SCM},i} > 0,
\end{equation}
where $\hat{p}_{\mathrm{c,SCM},i}$ is the observed SCM crash probability for the $i$-th seed case, estimated from the 2,000 simulation runs.

Rearranging (\ref{eq:assumpation}) yields the weight for re-simulated crashes:
\begin{equation} \label{eq:resim_weight}
    \omega_{\mathrm{sim,}i} \propto \frac{\omega_i}{n_{\mathrm{sim,}i} \cdot \hat{p}_{\mathrm{c,SCM},i}} = \frac{\omega_i}{n_{\mathrm{sim,c,}i}} \quad \text{for } \hat{p}_{\mathrm{c,SCM},i} > 0,
\end{equation}
where $n_{\mathrm{sim,c,}i}$ is the number of re-simulated crashes originating from the $i$-th seed case.

Finally, the weights for all re-simulated crashes were computed using (\ref{eq:resim_weight}) and scaled so that their sum equals the total number of collected crashes (7,888).
The resulting weighted dataset of 7,888 scenarios, hereafter referred to as the ``SCM-based'' dataset, is compared with the reference dataset.

It is important to note that the SCM-based dataset represents only a finite sampling of the SCM’s stochastic behavior, which introduces several limitations.
First, seed cases that yield zero re-simulated crashes are effectively treated as having zero crash probability and are therefore excluded, even though the absence of observed crashes may simply be a finite-sample effect.
Second, for seed cases with one or more observed crashes, the weighting procedure relies on the strong assumption that the empirically observed SCM crash probability is a reasonable approximation of the corresponding real-world crash probability.
As a consequence, both the set of included seed cases and their associated weights are sensitive to the number of simulation runs and may change with additional re-simulations.
While increasing the number of runs would yield more stable and robust estimates of SCM-induced pre-crash behavior, the total number of re-simulations was constrained by the substantial computational effort.
Given these limitations and because the primary aim is to illustrate the proposed equivalence-testing framework, the SCM-based dataset is used here primarily for demonstration purposes, rather than to assess the validity of the SCM for crash generation.

\subsection{Metrics}
In this demonstration, the general scope is to assess the impact of an AEB system on Maximum Abbreviated Injury Scale (MAIS)~2+ injuries \citep{gennarelli2006ais} in rear-end pre-crash scenarios.
Following our previous study~\citep{wu2025practical}, we selected three of the previously used metrics and added one new metric, replacing the lead vehicle’s Delta-v with its (estimated) injury risk ($P_{\mathrm{inj}}$).
They provide a balanced basis for comparing reference/real-world and synthetic pre-crash scenario datasets in the safety impact assessment context.
\begin{itemize}
    \item $P_{\mathrm{inj}}$: Estimated MAIS 2+ injury risk for the lead vehicle's driver.
    This metric represents the severity of potential crash outcomes and serves as a severity-weighted measure for comparing scenario datasets.
    \item $t_{\mathrm{nr}}$ [$\mathrm{s}$]: No-return time.
    It is defined as the point of no return beyond which a collision is unavoidable even if the following vehicle applies the maximum deceleration of $-9~\mathrm{m/s^2}$.
    Time zero corresponds to the impact moment; thus, $t_{\mathrm{nr}}$ is negative.
    This metric measures scenario criticality and the temporal margin available for evasive action.
    \item $a_{\mathrm{l,min}}$ [$\mathrm{m/s^2}$]: Minimum acceleration (maximum deceleration) of the lead vehicle.
    This value characterizes the harshness of the lead vehicle’s braking maneuver, a common contributing factor in rear-end crashes.
    Capturing its distribution helps ensure that synthetic scenarios, to some extent, reflect realistic lead-vehicle deceleration behavior as it is relevant to AEB performance.
    \item $a_{\mathrm{f,min}}$ [$\mathrm{m/s^2}$]: Minimum acceleration of the following vehicle.
    Capturing its distribution helps ensure that synthetic scenarios reflect, to some extent, the realistic deceleration behavior of the following vehicle.
\end{itemize}

The four metrics collectively capture outcome severity ($P_\mathrm{inj}$), scenario criticality ($t_\mathrm{nr}$), and braking responses for the leading and following vehicles  ($a_\mathrm{l,min}$ and $a_\mathrm{f,min}$, respectively).
With this selection, practical equivalence testing considers factors that are physically meaningful and relevant to the assessment scope for which the synthetic scenarios are intended.
Note again that this is an example and that specific assessment scopes may need to capture other features of the pre-crash kinematics.

\subsection{Bin-Weight Function} \label{section:bin_weight}
An empirical approach for designing bin weights is described in Section~\ref{section:bin_weight_design}.
In this approach, the pre-crash scenarios in the reference dataset are first re-simulated with a specific DAS system (e.g., AEB in this work).
A set of re-simulation outcome measures most relevant to the assessment is then selected and used to define a bin-weight function based on prioritization judgment and domain knowledge of the assessment scope.

In the context of safety impact assessments, relevant outcome measures may include the re-simulated crash rate, the average re-simulated injury risk of the lead‑vehicle driver, the reduced average re-simulated injury risk of the lead‑vehicle driver, or other comparable measures.

In this demonstration, the weight of the $i$-th bin is defined as:
\begin{equation} \label{eq:bin_weight}
    \omega_i = f(\bar{P}_{\mathrm{inj,rs},i}) = \frac{\bar{P}_{\mathrm{inj,rs},i} + \varepsilon}{P_0 + \varepsilon},
\end{equation}
where $\bar{P}_{\mathrm{inj,rs},i}$ denotes the average probability of the lead-vehicle driver sustaining a MAIS~2+ injury for the re-simulated scenarios in the $i$-th bin group, $P_0$ is a pre-defined baseline MAIS~2+ injury risk, and $\varepsilon$ is a small positive constant introduced to ensure that the minimum possible weight is above zero.
A hypothetical baseline bin, defined as having a standard weight $\omega_\mathrm{b} = 1$, corresponds to the baseline injury risk $P_0$ (i.e., $\omega_i = 1$ when $\bar{P}_{\mathrm{inj,rs},i} = P_0$).
Note that, in this study, the MAIS~2+ injury risk of the lead-vehicle driver in non-crash cases is set to zero, and in crash cases it is computed using the model proposed by Wang~\citep{wang2022mais}:
\begin{equation}
    P_{\mathrm{inj}} = \frac{1}{1 + e^{6.1818 - 0.3315\Delta v_\mathrm{l}}},
\end{equation}
where $\Delta v_\mathrm{l}$~($\mathrm{m/s}$) is the speed change of the lead vehicle during the impact, estimated as described in Section~III-F of our previous study~\citep{wu2025model}.

With this bin-weight function, higher weights are assigned to scenarios with higher average re-simulated injury risks, thereby placing a greater emphasis on conditions in which the AEB system is less effective.
\textcolor{black}{Conversely, scenarios associated with lower average injury risks receive smaller weights, reflecting their comparatively limited influence on the assessment.}
This ranking aligns with the objective of safety impact assessment: to prioritize the accurate representation of safety-critical scenarios that most influence the estimated overall safety benefit of the system.

We set $\varepsilon = 1 \times 10^{-4}$ and $P_0 = 0.02$.
The choice of $P_0$ is motivated by conventional practice in injury risk modeling; a MAIS~2+ probability below 2–5\% is regarded as the lower bound of clinically meaningful injury risk \citep{forman2012predicting}.
Anchoring at 2\% avoids over-weighting near-zero cases, while ensuring that bins with higher risks are emphasized relative to a conservative but nontrivial reference point.
This value represents what is generally considered a meaningful baseline \citep{forman2012predicting}.
Note that the baseline is primarily used to scale the bin weights; one can choose any baseline.
Yet, in general, a baseline with an interpretable physical meaning is recommended, as it is more straightforward to set realistic ROPE thresholds accordingly (see Section \ref{section:parameters}).

\subsection{Parameters} \label{section:parameters}
The four parameters $N$, $\alpha$, $\theta_\mathrm{thd}$, and $\Theta_\mathrm{thd}$, described in Section \ref{section:param_setting}, need to be specified beforehand.
Considering the small sample size $n_\mathrm{ref} = 200$ of the reference dataset, we set $m = 40$, thereby obtaining the number of bins $N = 5$.

The remaining three parameters jointly define the equivalence criteria.
In this demonstration, each metric has to satisfy the equivalence criteria individually in order to establish practical equivalence.
For each metric, the posterior probability threshold is set to $\alpha = 0.95$.
A common pair of ROPE thresholds ($\theta_\mathrm{thd}$, $\Theta_\mathrm{thd}$) is applied across all metrics, ensuring comparability of the equivalence decisions (see Section \ref{section:param_setting}).

We propose a practical rule for setting ROPEs: specify tolerances for a hypothetical baseline bin and derive ROPEs from them.
As mentioned in Section \ref{section:bin_weight}, the baseline bin is assigned a weight of $\omega_\mathrm{b} = 1$.
The user should first interpret what this baseline condition represents in practical terms according to the chosen bin-weight function, ensuring that the tolerance specified is meaningful for the specific assessment.
The next step is to decide the acceptable range of deviations.
The absolute relative deviation $\left| \Delta P / P_\mathrm{ref}\right|$ and the absolute deviation $\left| \Delta P\right|$ (based on the definitions of $\theta$ and $\Theta$) should be regarded as practically equivalent for the hypothetical baseline bin.
The tolerances are denoted as $\left| \Delta P / P_\mathrm{ref}\right|_\mathrm{thd}$ and $\left| \Delta P\right|_\mathrm{thd}$.

In our demonstration, the hypothetical baseline bin corresponds to an average baseline MAIS~2+ injury risk $P_0 = 0.02$.
We assigned tolerances of $\left| \Delta P / P_\mathrm{ref}\right|_\mathrm{thd} = 10\%$ and $\left| \Delta P\right|_\mathrm{thd} = 5\%$ to this bin.
\textcolor{black}{These values mean that for the baseline bin, its proportion in the synthetic distribution $P_\mathrm{syn}$ must satisfy the two conditions:
\begin{equation}
    0.9 P_\mathrm{ref} \leq P_\mathrm{syn} \leq 1.1 P_\mathrm{ref},
\end{equation}
\begin{equation}
    P_\mathrm{ref} -0.05 \leq P_\mathrm{syn} \leq P_\mathrm{ref} + 0.05.
\end{equation}
Differences within ±10\% relative deviation or ±5\% absolute deviation are considered practically negligible and too small to materially influence the intended safety impact assessment.}

Once these tolerances are defined, the corresponding ROPE thresholds for $\theta$ and $\Theta$ can be derived directly from (\ref{eq:theta}) and (\ref{eq:Theta}) as:
\begin{equation}
    \theta_{\mathrm{thd}} = \left| \Delta P / P_\mathrm{ref}\right|_\mathrm{thd} \cdot \omega_\mathrm{b} = \left| \Delta P / P_\mathrm{ref}\right|_\mathrm{thd},
\end{equation}
\begin{equation}
    \Theta_{\mathrm{thd}} = \left| \Delta P\right|_\mathrm{thd} \cdot \omega_\mathrm{b} = \left| \Delta P\right|_\mathrm{thd}.
\end{equation}
Based on the chosen tolerances, the ROPE thresholds are thus $\theta_\mathrm{thd} = 0.10$ and $\Theta_\mathrm{thd} = 0.05$.
A 10\% localized deviation tolerance ensures sensitivity to meaningful discrepancies in critical bins, while a 5\% aggregate tolerance maintains a conservative standard for overall equivalence.
This balance follows the general principle that ROPE boundaries should be defined based on domain-relevant effect sizes (that is, differences deemed practically meaningful for the assessment purpose) rather than on arbitrary statistical thresholds \citep{kruschke2018rejecting, schwaferts2020bayesian}.
In summary, this practical procedure offers an intuitive approach to translating user-specified practical tolerances into consistent formal ROPE thresholds that can be applied across bins.

\begin{figure}[!t]
    \centering
    \includegraphics[width=0.5\linewidth]{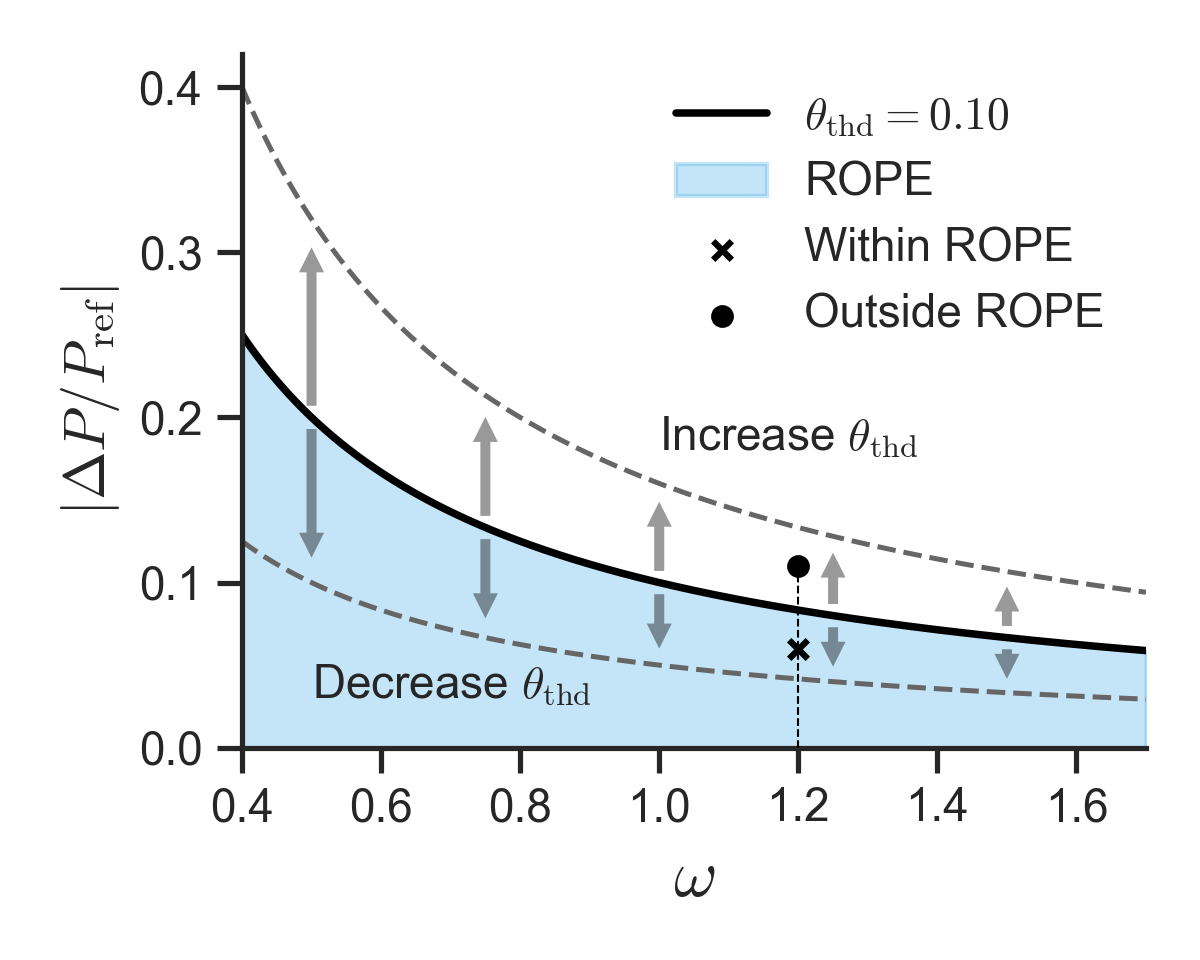}
    \caption{Illustration of the ROPE for $\theta$.
    The x-axis represents the weight of the bin with the maximum weighted absolute relative deviation, while the y-axis represents the absolute relative deviation.
    The cross marker at (1.2,~0.06) lies within the ROPE, whereas the circle marker at (1.2,~0.11) lies outside.
    The solid black curve corresponds to $\theta_\mathrm{thd}=0.10$, with the shaded region representing the ROPE}
    \label{fig:theta_ROPE}
\end{figure}

Fig.~\ref{fig:theta_ROPE} illustrates the defined ROPE for $\theta$.
The arrows and dashed gray curves demonstrate how variations in $\theta_\mathrm{thd}$ affect the ROPE: an increase in $\theta_\mathrm{thd}$ results in a larger ROPE, while a decrease results in a smaller ROPE.
At $\omega=1.2$, the threshold for $\left| \Delta P/P_\mathrm{ref}\right|$ is $\theta_\mathrm{thd}/\omega = 0.10 / 1.2 \approx 0.083$, indicating that the cross marker at (1.2,~0.06) lies within the ROPE, whereas the circle marker at (1.2,~0.11) lies outside.
This visualization illustrates that the equivalence decision is sensitive to the definition of tolerances, underscoring the importance of transparent justification when setting thresholds.

\subsection{Results}
\subsubsection{Weighted PCM dataset} \label{section:results_weighted_pcm}
\begin{figure}[!t]
    \centering
    \includegraphics[width=0.4\linewidth]{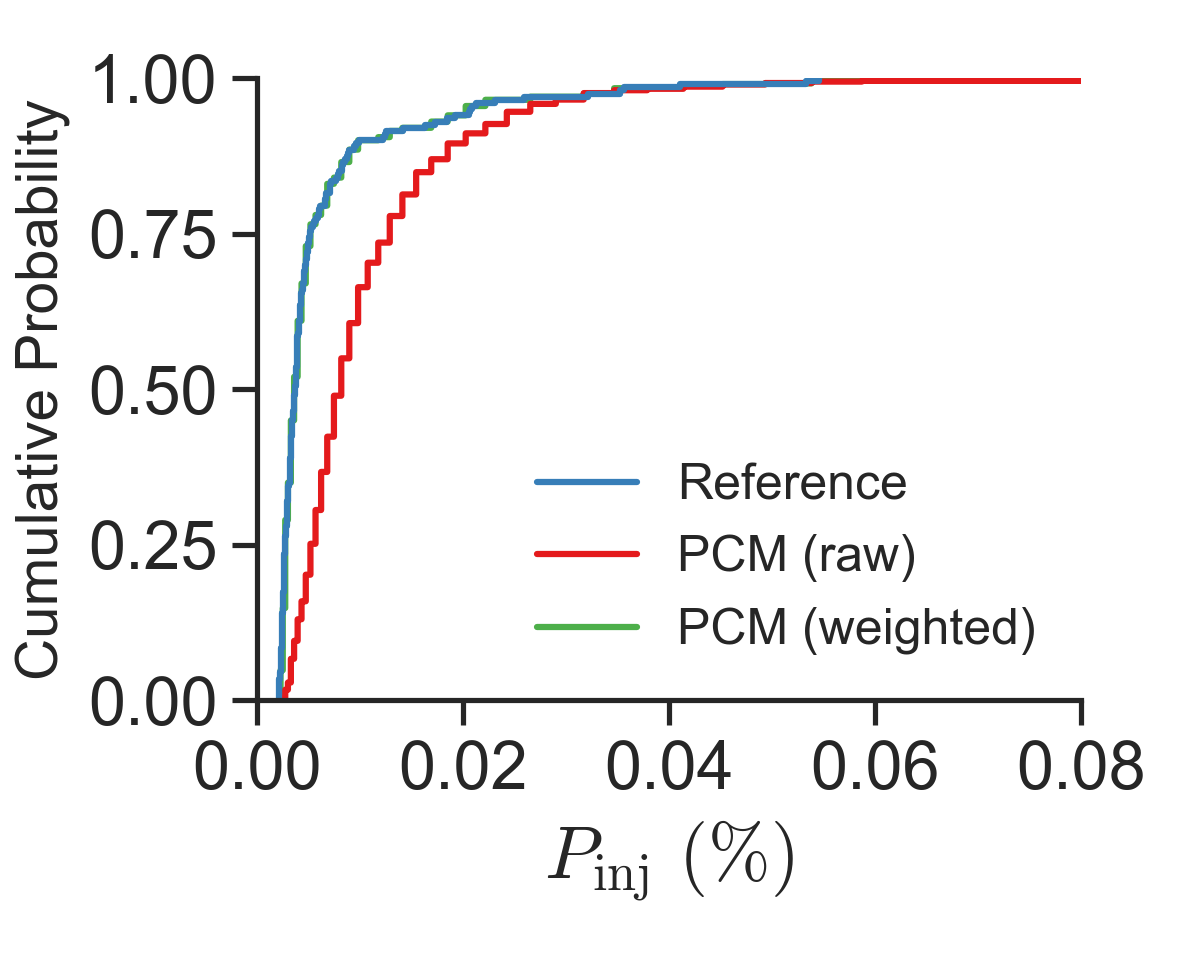}
    \caption{Cumulative distribution functions (CDFs) of the injury risk $P_\mathrm{inj}$ for the reference dataset (blue), the raw PCM dataset (red), and the weighted PCM dataset (green).
    The green curve closely overlaps with the blue one.}
    \label{fig:datasets}
\end{figure}

As noted in Section~\ref{section:datasets}, the PCM dataset includes only pre-crash data that resulted in the injury of at least one person.
To mitigate this limitation, we generated a weighted version of the dataset using the k-nearest neighbors (KNN)-based sample weighting method developed in our previous study~\citep{wu2025model}, aligning the injury risk distribution with that of the reference dataset.
Figure~\ref{fig:datasets} shows the cumulative distributions of $P_\mathrm{inj}$ for the reference, raw PCM, and weighted PCM datasets, illustrating how the weighting procedure effectively corrects for the sampling bias in this application.
Equivalence tests were then performed to compare both the raw and weighted PCM datasets against the reference dataset.

\subsubsection{Re-simulation}
\begin{figure}[!t]
    \centering
    \subfloat[]{\includegraphics[width=0.2\textwidth]{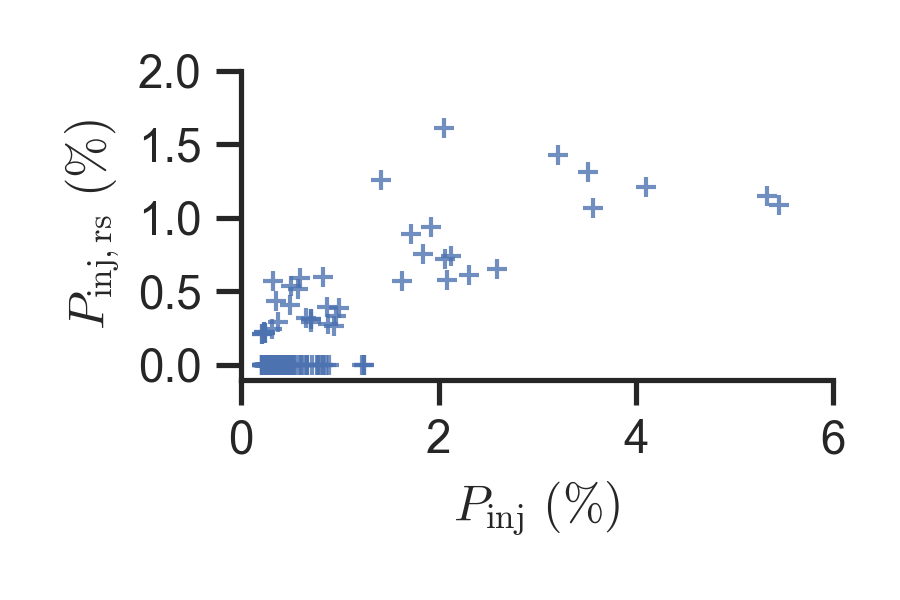}}
    \hfil
    \subfloat[]{\includegraphics[width=0.2\textwidth]{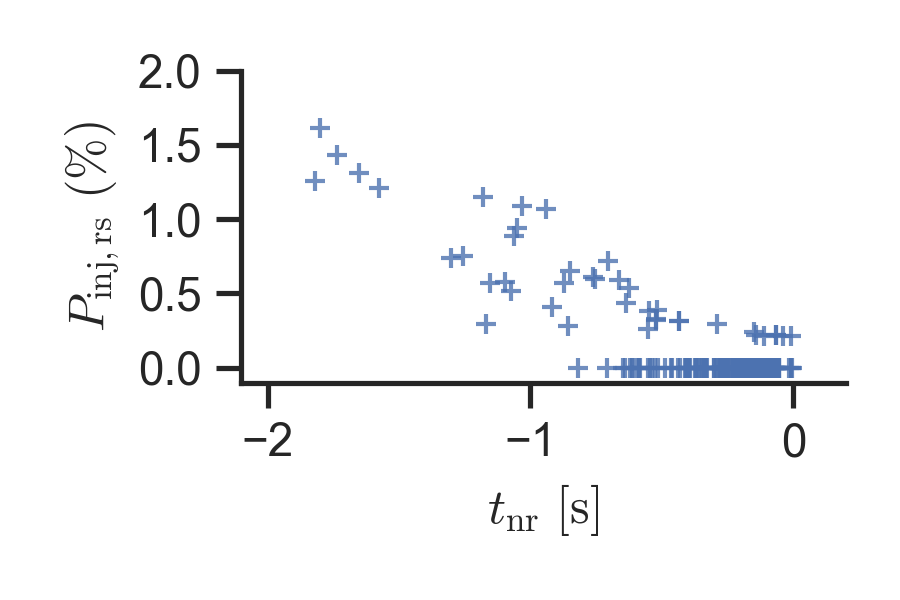}}
    \vfil
    \subfloat[]{\includegraphics[width=0.2\textwidth]{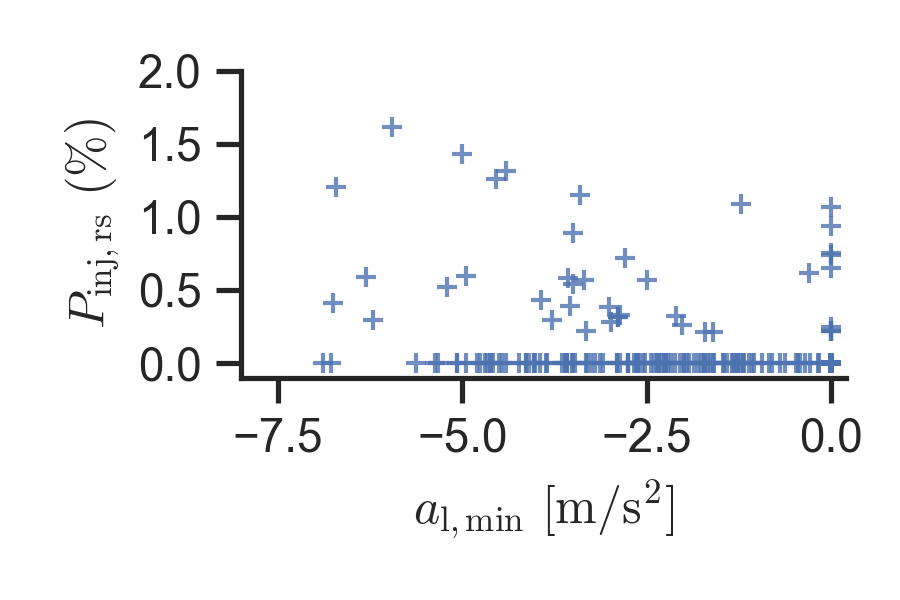}}
    \hfil
    \subfloat[]{\includegraphics[width=0.2\textwidth]{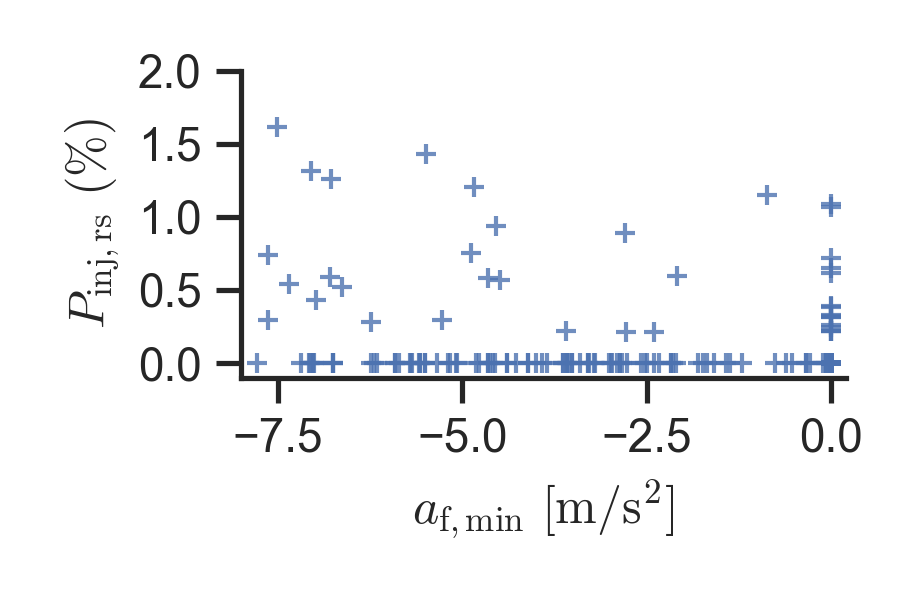}}
    \caption{Re-simulated injury risk of the lead-vehicle driver against four metrics in the reference dataset: (a) $P_\mathrm{inj}$, (b) $t_\mathrm{nr}$, (c) $a_{l,\min}$, and (d) $a_{f,\min}$.}
    \label{fig:resim_results}
\end{figure}

As described in Section \ref{section:bin_weight}, to calculate bin weights, the pre-crash scenarios in the reference dataset are re-simulated with the AEB system; of the 200 crashes, 41 remain.
Fig.~\ref{fig:resim_results} shows the re-simulation results: higher re-simulated injury risks are associated with greater baseline injury risk, shorter no-return times, and harsher braking maneuvers.
This information provides the basis for assigning larger bin weights to more critical scenarios in the equivalence tests.
The weights are used to determine the two statistics $\theta$ and $\Theta$ for each metric.

\subsubsection{Statistics computation and diagnostic insights}
\begin{figure}[!t]
    \centering
    \subfloat[]{\includegraphics[width=0.4\textwidth]{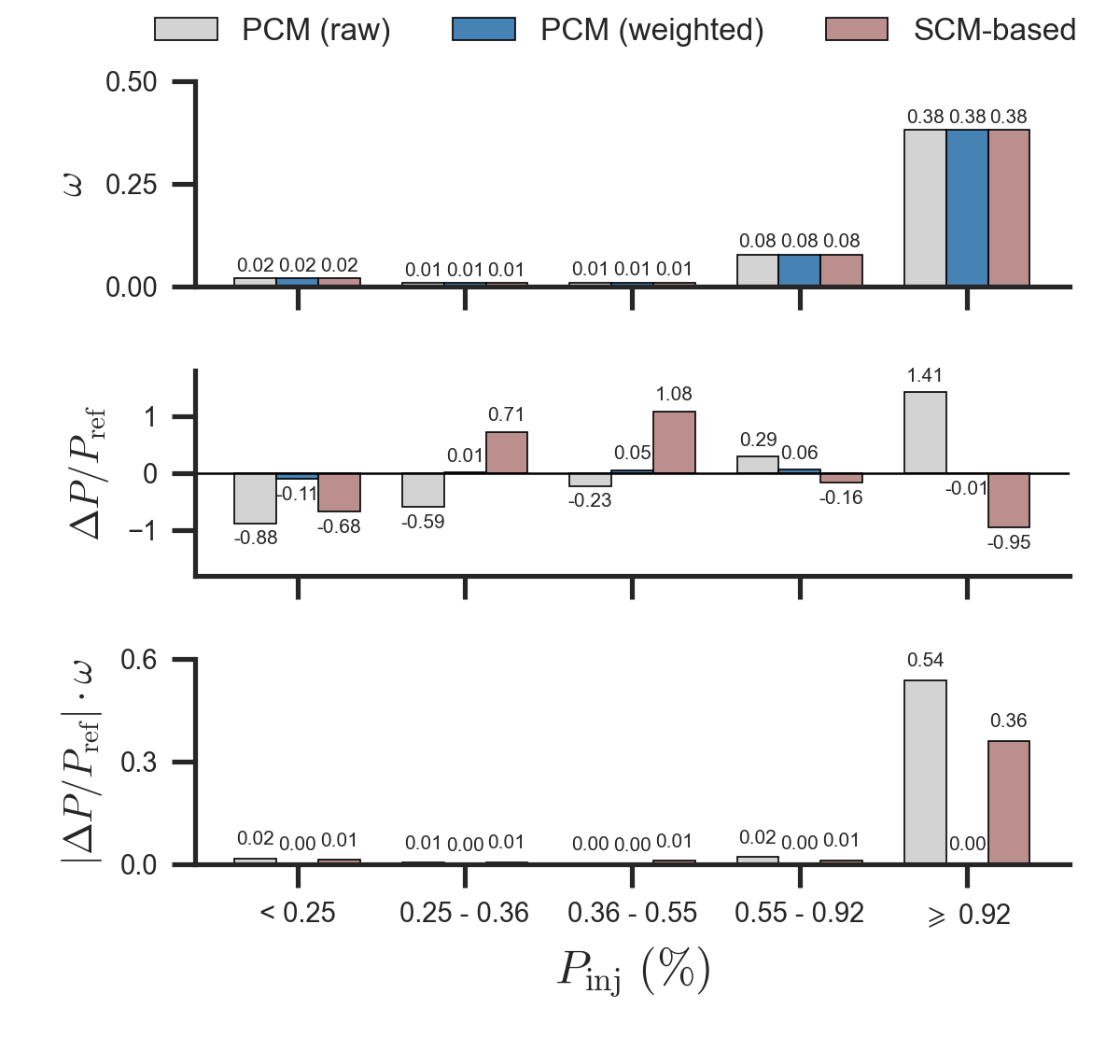}}
    \vfil
    \subfloat[]{\includegraphics[width=0.4\textwidth]{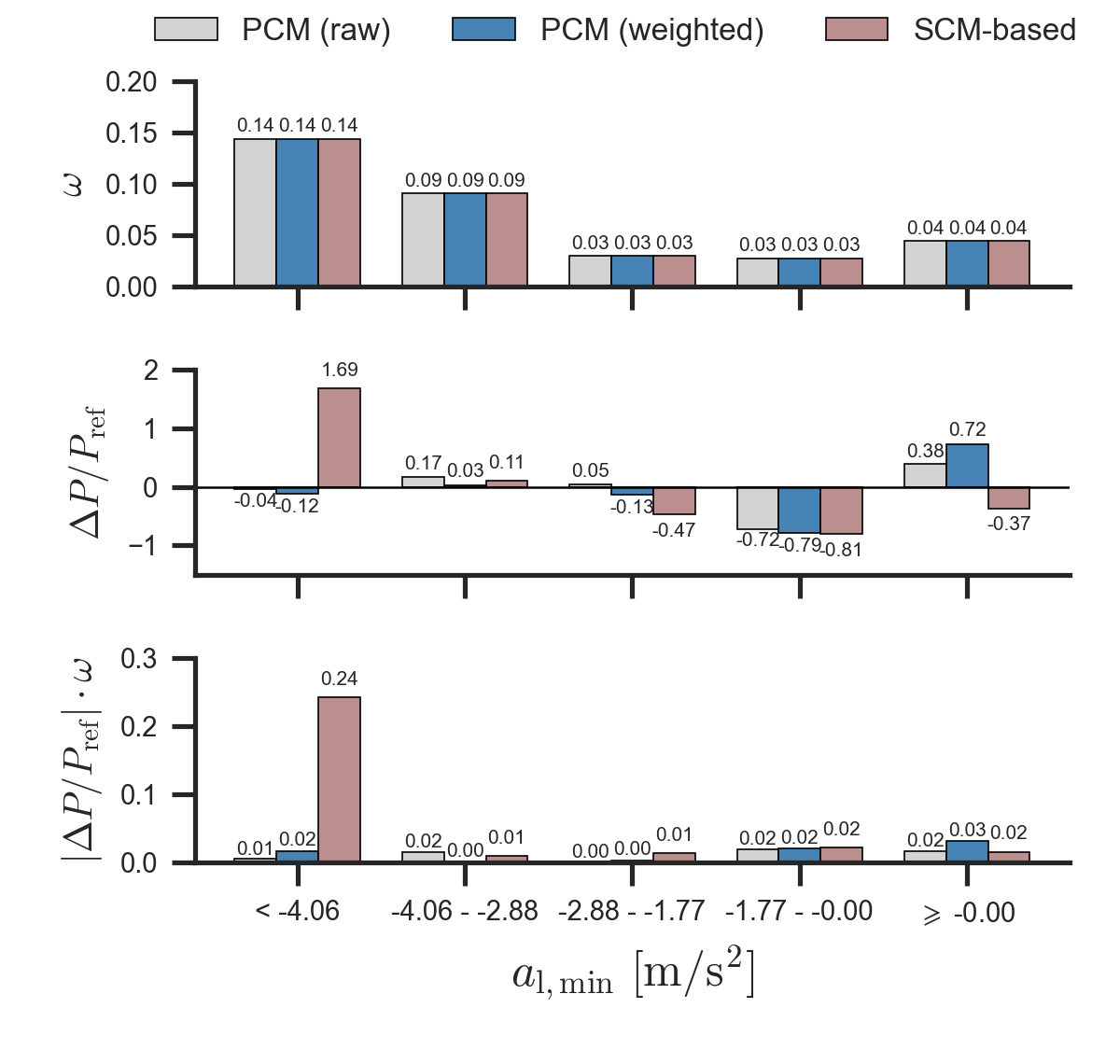}}
    \caption{Illustrative examples of $\theta$ for (a) the injury risk metric $P_\mathrm{inj}$ and (b) the lead vehicle’s minimum acceleration $a_{l,\min}$.
    For (a) and (b), the top figures show the bin weights $\omega$; the middle, the relative deviations $\Delta P/P_\mathrm{ref}$; and the bottom, the weighted absolute relative deviations $|\Delta P/P_\mathrm{ref}|\cdot \omega$ for one selected paired draw from the posterior distributions of the fitted models for the reference dataset and the dataset being compared.}
  \label{fig:theta_examples}
\end{figure}

\begin{table*}[!t]
\centering
\caption[l]{Statistics Across Metrics and Comparison Types}
\label{tab:equivalence_stats}
\begin{tabular}{lllllllll}
\toprule
\textbf{Metric} & \textbf{Statistic} & \textbf{ROPE} & \textbf{Raw PCM} & & \textbf{Weighted PCM} & & \textbf{SCM-based} &\\
& & & \textbf{95\% HDI} & \textbf{Equivalence} & \textbf{95\% HDI} & \textbf{Equivalence} & \textbf{95\% HDI} & \textbf{Equivalence}\\
\midrule
$P_\mathrm{inj}$ & $\theta$ & [0, 0.10] & [0.48, 0.58] & No & [0.00, 0.06] & Yes & [0.26, 0.41] & No\\
& $\Theta$ & [0,0.05] & [0.10, 0.12] & No & [0.00, 0.01] & Yes & [0.05, 0.09] & No\\
$a_\mathrm{l,min}$ & $\theta$ & [0, 0.10] & [0.02, 0.05] & Yes & [0.02, 0.05] & Yes & [0.22, 0.30] & No\\
& $\Theta$ & [0, 0.05] & [0.01, 0.02] & Yes & [0.01, 0.02] & Yes & [0.05, 0.07] & No\\
$a_\mathrm{f,min}$ & $\theta$ & [0, 0.10] & [0.07, 0.17] & No & [0.02, 0.09] & Yes & [0.21, 0.34] & No\\
& $\Theta$ & [0, 0.05] & [0.02, 0.04] & Yes & [0.01, 0.02] & Yes & [0.07, 0.08] & No\\
$t_\mathrm{nr}$ & $\theta$ & [0, 0.10] & [0.30, 0.39] & No & [0.00, 0.05] & Yes & [0.04, 0.20] & No\\
& $\Theta$ & [0, 0.05] & [0.07, 0.08] & No & [0.00, 0.01] & Yes & [0.01, 0.05] & Yes\\
\bottomrule
\end{tabular}
\end{table*}

As described in Section~\ref{section:method_steps}, Bayesian distribution models were fitted to the reference and synthetic datasets for each metric.
The two statistics $\theta$ and $\Theta$ were then computed for every paired draw from the posterior distributions of the optimal models.
Finally, the 95\% HDI of each statistic’s posterior distribution was compared with the corresponding ROPE.

Fig.~\ref{fig:theta_examples} provides an example of a single instance of $\theta$ for the injury risk metric $P_\mathrm{inj}$ and the lead vehicle’s minimum acceleration $a_{l,\min}$.
The bin weights are identical across the raw and weighted PCM datasets as well as the SCM-based dataset, since they are determined by the same draw from the posterior distributions of the fitted model for the reference data.
In this instance, the highest weights are assigned to the bins with the highest $P_\mathrm{inj}$ and the lowest $a_\mathrm{l,min}$, which reflect the scenarios in which the AEB system is least effective at mitigating crashes.

$\Delta P/P_\mathrm{ref}$ measures the relative deviation across bins between the paired distribution draws.
$|\Delta P/P_\mathrm{ref}|\cdot \omega$ then scales this deviation by the importance of the corresponding bin.
Finally, $\theta$ is defined as the maximum of the weighted deviations across all bins; it captures the most critical (localized) distributional discrepancy, considering its magnitude as well as its relevance to the intended assessment.
The computation of $\Theta$ follows the same binning process, except that instead of choosing the maximum weighted absolute relative deviation, it summarizes the weighted absolute deviations across all bins.

For diagnostic purposes, the weighted relative and absolute deviations of each bin ($|\Delta P_i/P_{\mathrm{ref},i}|\cdot \omega_i$ and $|\Delta P_i|\cdot \omega_i$, respectively) can also be recorded and analyzed.
The contribution of each bin to each statistic is useful information which can be utilized to enhance dataset generation or weighting strategies.
For example, it may lead to improved sampling procedures, or more refined generative models which better represent the distributional characteristics of those bins with a highly weighted divergence.

\textcolor{black}{For the specific case illustrated in Figure \ref{fig:theta_examples}a, for metric $P_\mathrm{inj}$, among the three datasets, the raw PCM dataset contains the highest proportion of high-severity scenarios than the reference dataset, resulting in $\theta = 0.54$.
Compared to the raw PCM dataset, the weighted PCM dataset aligns closely with the reference dataset, yielding a much smaller $\theta = 0.00$.
Meanwhile, the SCM-based dataset contains a lower proportion of high-severity scenarios, resulting in $\theta = 0.36$.}

\textcolor{black}{For the case illustrated in Figure \ref{fig:theta_examples}b, for metric $a_\mathrm{l,min}$, among the three datasets, the SCM-based dataset contains the highest proportion of scenarios with low $a_\mathrm{l,min}$ values than the reference dataset, resulting in $\theta = 0.24$.
This difference may be explained by the relatively small sample size; seed cases involving a harsh-braking lead vehicle are more likely to produce crash events even within the limited number of simulation runs, while crashes from seed cases with a lower crash probability may simply not appear due to finite-sample limitations.
Nevertheless, we cannot rule out the possibility that part of the observed discrepancy arises from limitations of the SCM itself rather than sampling effects alone.}
In contrast, the raw and weighted PCM datasets align closely with the reference dataset, yielding much smaller values of $\theta$.

\subsubsection{Equivalence testing results}
Table~\ref{tab:equivalence_stats} summarizes the results of the equivalence tests.
\begin{itemize}
    \item \textbf{Raw PCM dataset (non-equivalence)}: The statistics for $P_\mathrm{inj}$ and $t_\mathrm{nr}$ both exceeded the ROPE thresholds, indicating systematic deviations from the reference dataset.
    \textcolor{black}{In addition, $a_\mathrm{f,min}$ failed the equivalence test due to the large proportion of harsh-braking events in the PCM data, which shifts the distribution toward stronger deceleration values.
    These results reflect the expected bias of the raw PCM dataset toward more severe cases; compared with the reference dataset, the raw PCM dataset exhibits higher injury risks (see Figure \ref{fig:theta_examples}a), shorter no-return times, and stronger braking responses.}
    \item \textbf{Weighted PCM dataset (equivalence)}: In contrast, the weighted PCM dataset satisfied the equivalence criteria across all four metrics.
    This finding provides evidence that carefully designed weighting strategies can restore distributional comparability between biased and representative datasets.
    However, it is important to be aware that the result applies specifically to the application considered here; in general, weighting may not fully correct for structural biases, unobserved confounders, or differences in underlying data-generating mechanisms.
    The effectiveness of sample weighting should therefore be evaluated on a case-by-case basis.
    \item \textbf{SCM-based dataset (non-equivalence)}: None of the metrics satisfy the equivalence criteria.
    As shown in Fig.~\ref{fig:theta_examples}, the SCM-based dataset under-represents high-severity scenarios and over-represents cases involving a harsh-braking lead vehicle, perhaps due to the limited number of simulation runs per seed case.
\end{itemize}

\section{Discussion and Conclusions} \label{section:contri}

\subsection{Contributions}
\textcolor{black}{A critical gap exists in validating the representativeness of the synthetic scenarios used for safety impact assessment of DAS.
To ensure an accurate and credible assessment, it is essential to determine whether the synthetic scenarios are practically equivalent to their real-world counterparts for the intended assessment.
However, existing validation practices in traffic safety, such as conventional statistical significance tests, are inadequate for this purpose.}
To address this gap, this study proposes and demonstrates an extension of the practical equivalence testing framework for validating synthetic pre-crash scenarios to assess the safety impact of DAS, building on our earlier conference work \citep{wu2025practical}.
The extension introduces two binning-based statistics, $\theta$ and $\Theta$, which quantify localized and aggregate distributional discrepancies between datasets, accounting for both the magnitude of the differences and their practical relevance.
Practical relevance is incorporated by weighting the bins according to system-specific re-simulation outcomes, such as injury risk or crash rate, thereby emphasizing conditions that pose the greatest challenges to the DAS under assessment.

The case study yielded several insights.
First, the raw PCM dataset failed equivalence tests due to its inherent bias toward higher-severity crashes.
Second, applying a KNN-based reweighting procedure to that dataset resulted in equivalence across all metrics, demonstrating that appropriate weighting can mitigate bias in injury-focused datasets and align them with the reference dataset.
Third, the SCM-based dataset failed most equivalence tests.
\textcolor{black}{This result should be interpreted cautiously, as it may primarily reflect finite-sample effects arising from the limited number of simulation runs, in which many seed cases exhibited zero observed crashes and were thus treated as having zero probability.
Therefore, the present data do not permit a firm conclusion about the adequacy of the SCM; the apparent non-equivalence may be driven by insufficient sampling rather than model limitations.}
Finally,  the examples demonstrate that the proposed statistics offer diagnostic insight into which bin groups drive non-equivalence, enabling targeted identification of where synthetic datasets diverge most critically from the reference dataset.

Beyond the specific application demonstrated here, the framework contributes methodologically by offering a structured, transparent approach to validating synthetic datasets that goes beyond conventional significance testing.
By providing interpretable, relevance-weighted measures of practical difference, the framework supports both quantitative evaluation and diagnostic analysis in safety impact assessment.

\subsection{Practical implications and recommendations}
The proposed framework offers a structured workflow that practitioners can follow when validating synthetic datasets for safety impact assessment.
By specifying assessment-relevant metrics and applying the binning-based statistics $\theta$ and $\Theta$, users can obtain interpretable, decision-relevant measures with practical equivalence between synthetic and reference data.
The two statistics offer complementary perspectives, enabling both localized and aggregate assessment of deviations across a wide range of validation contexts: $\theta$ highlights worst-case localized discrepancies, while $\Theta$ captures aggregate distributional differences. 
Importantly, because they operate on distributional representations of user-defined metrics rather than on scenario-specific assumptions, the two statistics are generic and agnostic to the specific application domain.

\textcolor{black}{Before conducting formal equivalence testing, however, it is essential to perform basic sanity and plausibility checks on the synthetic scenarios.
Sanity checks verify that key aggregate statistics fall within reasonable ranges, such as overall crash rates, proportions of crash types, and other high-level characteristics of the scenario set.
These checks help detect implementation errors or unintended biases in the scenario generation process.}

Plausibility checks then ensure the synthetic scenarios are physically and behaviorally reasonable, independent of any comparison with reference data.
For example, basic kinematic quantities such as speed, acceleration, and jerk should fall within physically feasible and behaviorally realistic ranges.
Scenarios that violate these fundamental constraints should be identified and addressed prior to equivalence testing, as they reflect modeling or simulation artifacts rather than issues of representativeness.

Once these sanity and plausibility checks are satisfied, equivalence testing should be used to assess representativeness relative to a reference dataset.
At this stage, users should consider whether key scenario types are adequately represented and whether known dataset limitations, such as finite simulation runs or uncorrected sampling biases, may lead to non-equivalent outcomes.
In such cases, failing an equivalence test does not necessarily indicate deficiencies in the underlying behavioral or system models; instead, it may reflect limitations in the data generation or sampling process (see the SCM-based dataset as an example).

\textcolor{black}{It is also important to note that, although appropriate weighting is demonstrated to be effective in this work, it cannot compensate for all structural differences.
If the underlying data-generating mechanisms differ fundamentally, or if relevant dimensions are unobserved, reweighting may only partially mitigate bias.
Therefore, the effectiveness of weighting should be evaluated on a case-by-case basis using assessment-oriented diagnostics, such as the proposed bin-level statistics.}

\textcolor{black}{Additionally, in practice, comprehensive reference datasets covering the entire joint parameter space are rarely available.
Instead, only subsets of parameters are typically found within available reference datasets.
Under such conditions, it is still possible to conduct equivalence testing.
However, equivalence criteria cannot be inferred solely from population structure; instead, they must be specified using external inputs, such as expert judgment, prior evidence, sensitivity analyses, or assessment requirements linked to system design or regulatory context.
Consequently, validation results should be interpreted as conditional on these assumptions, rather than as definitive statements about population-level representativeness.
Explicitly documenting these assumptions is essential for transparency, reproducibility, and meaningful use of validation results in safety assessment and regulatory dialogue.}

Finally, equivalence testing should be viewed as an iterative diagnostic tool rather than a one-off pass/fail criterion in the context of safety impact assessment.
When non-equivalence is detected, the binning-based statistics provide insight into which parts of the distribution drive the discrepancy, guiding targeted improvements to scenario generation, weighting strategies, and/or simulation design.
In this way, the framework supports the systematic refinement of synthetic datasets and their progressive alignment with assessment-relevant reference behavior.

\section{Limitations and Future Work}
Several limitations of the proposed framework should be acknowledged.
First, the framework does not yet provide practical guidelines for selecting the most relevant metrics for a given safety impact assessment (i.e., Step 1 in Section \ref{section:method_steps_1}).
Future work should establish systematic criteria or decision rules for metric selection to ensure that equivalence testing focuses on the metrics most relevant to the scope of the intended assessment.

Second, the current implementation treats metrics independently and therefore does not account for potential correlations among them, which may result in overly conservative practical equivalence decisions and a higher incidence of false non-equivalence conclusions \citep{wu2025practical}.
Expanding the framework to include multivariate distribution modeling could facilitate a more comprehensive and accurate evaluation of scenario realism.

Finally, the method is sensitive to prior assumptions of distribution models and the specification of ROPEs, which influence posterior inferences about equivalence.
Sensitivity analyses of prior choices and ROPE definitions will be essential for improving transparency.

\section*{Acknowledgments}
This research was supported by the Fordonsstrategisk forskning och innovation (FFI) program, sponsored by Vinnova, the Swedish governmental agency for innovation, as part of the project Improved quantitative driver behavior models and safety assessment methods for ADAS and AD (QUADRIS: nr. 2020-05156).
The authors wish to thank Mikael Ljung Aust at Volvo Cars Safety Center for reviewing the manuscript.

\section*{Code availability}
\textcolor{black}{The core implementation of the proposed equivalence testing procedure in this study is publicly available in the \texttt{bayes-binned-equivalence} repository \citep{Wu_bayes-binned-equivalence_2026}.}


\appendix
\section{Bootstrap-Based Power Analysis}
To assess the reliability and sensitivity of the proposed ROPE-based equivalence testing framework, a bootstrap analysis was conducted to estimate its power, defined as the probability of correctly declaring equivalence when the true parameter lies within the ROPE \citep{schwaferts2020bayesian}.
The bootstrap analysis evaluates how consistently the method identifies equivalence when it is indeed true, given the adopted parameter settings ($N$, $\alpha$, $\theta_\mathrm{thd}$, and $\Theta_\mathrm{thd}$) and sample sizes.

The same reference dataset used in the main analysis was used.
As described in Section \ref{section:datasets}, the dataset was randomly sampled from a large parent dataset, the QUADRIS pre-crash dataset.
One thousand bootstrap replicates (essentially new ``synthetic'' datasets) were created from the parent dataset, each the same size as the PCM dataset, using the same process that created the original reference dataset.
For each new synthetic dataset, the equivalence testing procedure was applied in the same manner as described in the main analysis.

Since those new datasets are truly equivalent to the reference dataset, power was estimated as the proportion of bootstrap replicates in which equivalence was declared.
Binomial 95\% confidence intervals for the estimated power were computed using the Wilson method \citep{wilson1927probable}.

\begin{table}[!t]
    \centering
    \caption[l]{Estimated power of the ROPE-based equivalence test} \label{tab:power}
    \begin{tabular}{llll}
    \toprule
    \textbf{Metric} & \textbf{Statistic} & \textbf{Power} & \textbf{95\% CI}\\
    \midrule
    $P_\mathrm{inj}$ & $\theta$ & 0.870 & [0.848, 0.889]\\
    & $\Theta$ & 1.000 & [0.996, 1.000]\\
    $a_\mathrm{l,min}$ & $\theta$ & 1.000 & [0.996, 1.000]\\
    & $\Theta$ & 1.000 & [0.996, 1.000]\\
    $a_\mathrm{f,min}$ & $\theta$ & 1.000 & [0.996, 1.000]\\
    & $\Theta$ & 1.000 & [0.996, 1.000]\\
    $t_\mathrm{nr}$ & $\theta$ & 0.997 & [0.991, 0.999]\\
    & $\Theta$ & 1.000 & [0.996, 1.000]\\
    \bottomrule
    \end{tabular}
\end{table}

The resulting power estimates are summarized in Table~\ref{tab:power}.
Across all metrics and both statistics ($\theta$ and $\Theta$), the estimated power exceeds 0.8 and, in most cases, approaches 1.0, indicating that the framework reliably detects practical equivalence under the adopted parameter setting.

\bibliographystyle{elsarticle-harv} 
\bibliography{references}

\end{document}